\documentclass[11pt,a4paper]{article}
\usepackage[hyperref]{emnlp2020}
\usepackage{times}
\usepackage{latexsym}
\usepackage{amsmath}
\usepackage{amssymb}
\usepackage{siunitx}  
\usepackage[ruled,linesnumbered]{algorithm2e}
\usepackage{url}
\usepackage{mathtools}

\usepackage{tabularx}
\usepackage{xspace}
\usepackage{booktabs}
\usepackage{collcell}

\usepackage{xfrac}

\usepackage{microtype}

\aclfinalcopy 
\def\aclpaperid{2337} 

\newcommand{\man}{\textsc{Man}\xspace}
\newcommand{\smart}{\textsc{Sem}\xspace}
\newcommand{\cdm}{c_0^{D}}
\newcommand{\csm}{c_0^{S}}
\newcommand{\cds}{c_1^{D}}
\newcommand{\css}{c_1^{S}}
\newcommand{\nop}[1]{}
\usepackage{soul}

\title{Learning a Cost-Effective Annotation Policy for Question Answering}

\date{\today}
\author{Bernhard Kratzwald$^{\diamondsuit\,\spadesuit}$ \quad \quad Stefan Feuerriegel$^\diamondsuit$ \quad \quad Huan Sun$^\spadesuit$ 
\\
        $^\diamondsuit$ Chair of Management Information Systems, ETH Zurich \\
        $^\spadesuit$ Department of Computer Science and Engineering, The Ohio State University \\
        {\tt \{bkratzwald, sfeuerriegel\}@ethz.ch \ sun.397@osu.edu} \\
}

\begin{document}
\maketitle

\begin{abstract}

State-of-the-art question answering (QA) relies upon large amounts of training data for which labeling is time consuming and thus expensive. For this reason, customizing QA systems is challenging. As a remedy, we propose a novel framework for annotating QA datasets that entails learning a cost-effective annotation policy and a semi-supervised annotation scheme. The latter reduces the human effort: it leverages the underlying QA system to suggest potential candidate annotations. Human annotators then simply provide binary feedback on these candidates. Our system is designed such that past annotations continuously improve the future performance and thus overall annotation cost. To the best of our knowledge, this is the first paper to address the problem of annotating questions with minimal annotation cost. We compare our framework against traditional manual annotations in an extensive set of experiments. We find that our approach can reduce up to 21.1\% of the annotation cost. 
 
\end{abstract}

\section{Introduction}
\begin{figure*}[t]
    \centering
    \includegraphics[width=.9\linewidth]{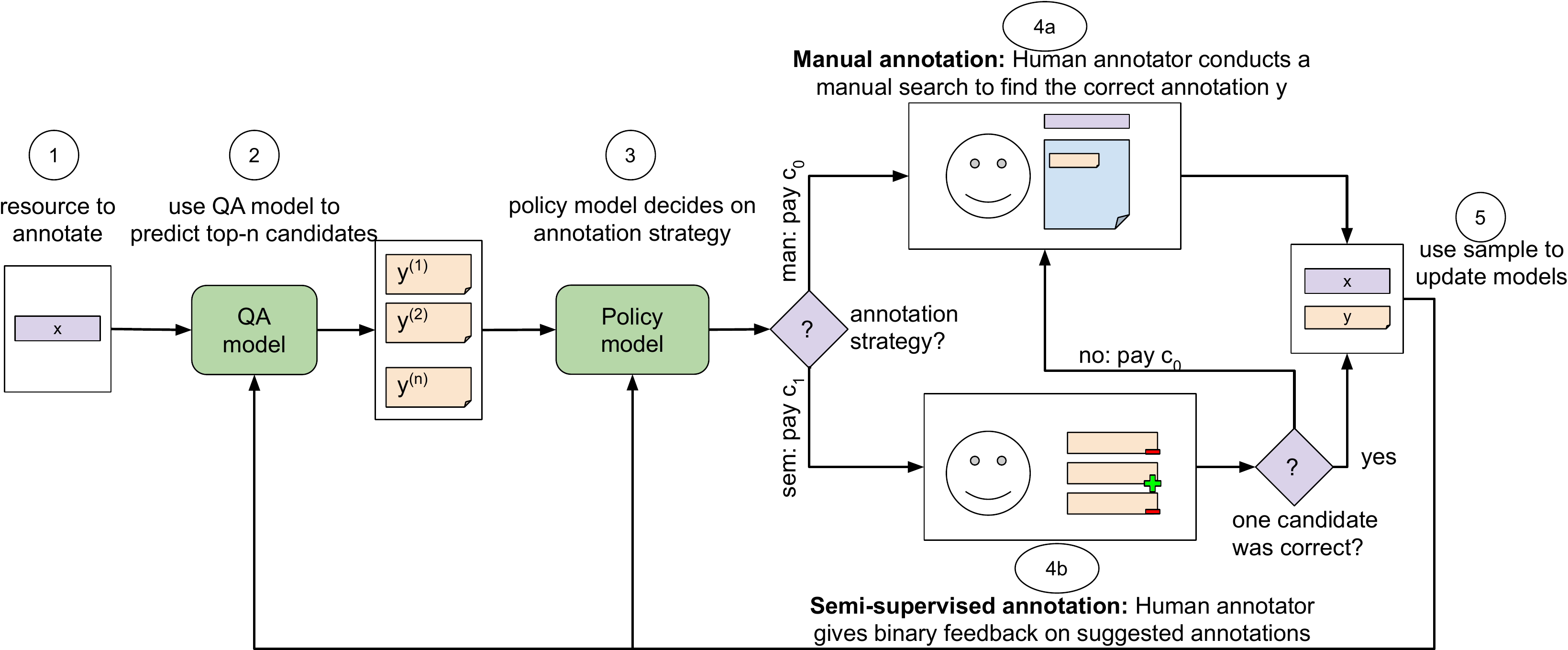}
    \caption{High-level overview of our framework: We leverage a QA model to predict candidate annotations for a given resource (e.g., $x$ stands for a question or question-document pair, while $y$ is the document or answer span). A policy model decides upon whether to invoke a \man or \smart scheme based on those predictions. {In the event that the semi-supervised strategy fails, we switch back to a manual annotation scheme. Finally, we use the annotated sample to update both the QA model and the policy model.}}
    \label{fig:overview_framwork}
\end{figure*}

Question answering (QA) based on textual content has attracted a great deal of attention in recent years \cite{Chen.2017,Lee.2018,Lee.2019,Yuqing.2020}. In order for state-of-the-art QA models to succeed in real applications (e.g., customer service), there is often a need for large amounts of training data. However, manually annotating such data can be extremely costly. For example, in many realistic scenarios, there exists a list of questions from real users (e.g., search logs, FAQs, service-desk interactions). Yet, annotating such questions is highly expensive \cite{Nguyen.2016,He.2018,Kwiatkowski.2019}: it requires the screening of a text corpus to find the relevant document(s) and subsequently screening the document(s) to identify the answering text span(s).

Motivated by the above scenarios, we study cost-effective annotation for question answering, whereby we aim to \emph{accurately}\footnote{By ``accurate,'' we mean that the resulting annotations will be of a similar quality to those from conventional manual annotation.} annotate a given set of user questions \emph{with as little cost as possible}. Generally speaking, there has been extensive research on how to reduce effort in the process of data labeling \cite{Haffari.2009}. For example, active learning for a variety of machine learning and NLP tasks \cite{Siddhant.2018} aims to select a small, yet highly informative, subset of samples to be annotated. The selection of such samples is usually coupled with a particular model, and thus, the annotated samples may not necessarily help to improve a different model \cite{Lowell.2019}. In contrast, we aim to annotate \emph{all} given samples at low cost and in a manner that can subsequently be used to develop any advanced model. This is particularly relevant in the current era, where a dataset often outlives a particular model \cite{Lowell.2019}. Moreover, there has also been some research into learning from distant supervision \cite{Yuqing.2020} or self-supervision~\cite{Sun.2019}. Despite being economical, such approaches often produce inaccurate or noisy annotations. In this work, we seek to reduce annotation costs without compromising the resulting dataset quality. 

We propose a novel annotation framework which learns a cost-effective policy for choosing between different annotation schemes, namely the conventional manual annotation scheme (\man) and a semi-supervised annotation scheme (\smart). Unlike the manual scheme, \smart does not require humans to screen a text corpus or document(s) in order to retrieve annotations. Instead, it leverages an initialized QA system, which can predict top-$n$ candidate annotations for documents or answer spans and asks humans to provide binary feedback (e.g., correct or incorrect) to the candidates. While this annotation scheme comes at a low cost, it fails when human annotators mark all candidates as incorrect. In such cases, the annotation cost has already been incurred and cannot be recouped. In order to produce an annotation, one must then draw upon the manual scheme (see Fig.~\ref{fig:overview_framwork}), in which case the policy would have been more effective if it had chosen the manual annotation scheme instead.
\textit{Therefore, how to choose the best annotation scheme for each question is the challenge we must address for this task.}

To tackle the above challenge, we propose a novel approach for learning a cost-effective policy. Here the policy receives several candidates and decides on this basis which annotation scheme to invoke. We train the policy with a supervised objective and learn a cost-sensitive decision threshold. The inherent advantage of this method is that our policy immediately reacts to changing costs (without re-optimizing model parameters) and does not exceed the cost of conventional manual annotation. Our policy is updated iteratively as more annotations are obtained.

We compare our framework against conventional, manual annotations in an extensive set of experiments. We simulate the annotation of NaturalQuestions~\cite{Kwiatkowski.2019}, as it consists of real user questions from search logs. Models in our framework are initialized with an existing dataset~\cite[SQuAD,][]{Rajpurkar.2016} and, as more annotations on NaturalQuestions become available, the framework is continuously updated. We study the sensitivity of our framework to varying cost ratios between \smart and \man. Our framework outperforms traditional manual annotation, even under conservative cost estimates for \smart, and in general reduces annotation costs in the range of 4.1\% to 21.1\%.

{All source code is publicly available from \href{https://github.com/bernhard2202/qa-annotation}{github.com/bernhard2202/qa-annotation}.}

\section{Related Work}
{\textbf{Question answering:}} In this paper, we study cost-effective annotation for question answering over textual content. There have been extensive efforts to create large-scale datasets for text-based QA, which have facilitated the development of state-of-the-art neural network based models~\cite[e.g.,][]{Chen.2017, Min.2018,Lee.2018,Kratzwald.2018,Wang.2018,Yuqing.2020}. Here we divide such datasets into two categories according to the way they were created: (1)~Datasets whose questions were created by crowdsourcing during the annotation process. Prominent examples include the Stanford Question and Answer Dataset \cite[SQuAD;][]{Rajpurkar.2016}, \mbox{HotPotQA} \cite{Yang.2018}, or NewsQA \cite{Trischler.2027}. (2)~``Natural'' datasets in which real-world questions are a~priori given. Here questions originate from, e.g., search logs or customer interactions. Prominent examples in this category include MS~MARCO \cite{Nguyen.2016}, DuReader \cite{He.2018}, or NaturalQuestions \cite{Kwiatkowski.2019}. This paper focuses on the latter category, that is, annotating ``natural'' datasets in a more cost-effective fashion where a set of questions is given. 

{\textbf{Active Learning:}} In the fields of machine learning and NLP, extensive research has been conducted on ways to reduce labeling effort \citep[e.g.,][]{Zhu.2008}. For example, the objective of active learning is to select only a small subset that is highly informative \cite[e.g.,][]{Haffari.2009} for annotation. To this end, researchers have developed various techniques based on, e.g., model uncertainty~\cite[cf.][]{Siddhant.2018}, expected model change~\cite{Cai.2013}, or functions learned directly from data \cite[e.g.,][]{Fang.2017}. However, the success of active learning is often coupled with a particular model and domain~\cite{Lowell.2019}. For instance, a dataset actively acquired with the help of an SVM model might underperform when used to develop an LSTM model. These problems become even more salient when complex black-box models are used in NLP tasks~\cite[cf.][]{chang2019overcoming}. {To summarize, active learning reduces annotation costs by deciding \emph{which} samples should be annotated. In our approach, we aim to annotate \emph{all} samples and study \emph{how} we should annotate them in order to reduce costs. Thus, the two approaches are orthogonal and can be combined.}

{\textbf{Learning from weak supervision and user feedback:}} Another approach to reducing annotation costs is changing full supervision to some form of weak (but potentially noisier) supervision. This has been adopted for various tasks such as machine translation \cite{Saluja.2012, Petrushkov.2018, Clark.2018, Kreutzer.2019}, semantic parsing \cite{Clarke.2010,Liang.2017,Talmor.2018}, or interactive systems that learn from user interactions~\cite{Iyer.2017, Gur.2018, Yao.2019, yao2020imitation}. For instance, \citet{Iyer.2017} used users to flag incorrect SQL queries. In contrast, similar approaches for text-based question answering are scarce. \citet{Joshi.2017} used noisy distant supervision to annotate the answer span and document for given trivia questions and their answers. \citet{Kratzwald.2019a} designed a QA system that continuously learns from noisy user feedback after deployment. In contrast to these works, this paper studies the problem of reducing labeling cost while maintaining accurate annotations.  

{\textbf{Quality estimation and answer triggering:} In a broader sense, this work is related to the literature on translation quality estimation \cite[e.g.,][]{martins.2017,specia.2013}. The goal in such works is to estimate (and possibly improve) the quality of translated text. Similarly, in question answering researchers use means of quality estimation for answer triggering~\cite{Zhao.2017,Kamath.2020}. Here, QA systems are given the additional option to abstain from answering a question when the best prediction is believed to be wrong. In our work, we estimate the quality of a set of suggested label candidates and, on the basis of these estimates we decide which annotation scheme to invoke.}
\section{Proposed Annotation Framework}

We study the problem of reducing the overall cost for annotating every given question $[q_1, \ldots, q_m]$. Specifically, our objective is to obtain the corresponding question-document-answer triples $\langle q_i,d_i,s_i \rangle$. In this paper, the natural language question $q_i$ is given, while we want to obtain the following annotations: the document from a text corpus  $d_i \in \mathcal{D}$ that contains the answer and the correct answer span $s_i$ within the document $d_i$. 

\subsection{Framework Overview}

Fig.~\ref{fig:overview_framwork} provides an overview of our framework for a cost-effective annotation of QA datasets. The framework comprises two main components: a \textbf{QA model} is used to suggest candidates for a resource to annotate while a \textbf{policy model} decides which annotation scheme to invoke (i.e., action). Our framework makes use of two annotation schemes:  a traditional \textbf{manual annotation scheme (\man)} and our \textbf{semi-supervised annotation scheme (\smart)}. Both annotation schemes incur different costs and, hence, the learning task is to find and update a cost-effective policy $\pi$ for making that decision.

\textbf{QA model:}
We define $\Omega$ as an arbitrary QA model over a text corpus $\mathcal{D}$ with the following properties. First, the model can be trained from annotated data samples, e.g., \mbox{$\Omega \leftarrow \text{train}\left(\{ \langle q_i,d_i,s_i \rangle\}_{0<i<\ldots}\right)$}. Second, for a given question the model can predict a number of top-$n$ documents likely to contain the answer, i.e., $\Omega^{D} : q \rightarrow [d^{(1)},\ldots, d^{(n)}] \in \mathcal{D}$. Third, for a given question-document pair the model can predict a number of top-$n$ answer spans, i.e., $\Omega^{S} : \langle q,d\rangle \rightarrow [s^{(1)},\ldots, s^{(n)}]$. These properties are fulfilled by recent QA systems dealing with textual content \cite[e.g.,][]{Chen.2017,Wang.2018}.

\textbf{Policy model:} For every question, we distinguish two policy models: a policy model $\pi^D$ responsible for annotating documents and $\pi^S$ for answer spans. For brevity, we sometimes drop the superscripts $S$ and $D$ and simply refer to them as $\pi$. The policy models decide whether a manual annotation scheme or rather our proposed semi-supervised annotation scheme is used, each of which is associated to different costs. 

\subsection{Annotation Schemes}

\textbf{Manual annotation (\man) scheme:} 
This scheme represents the status quo in which all annotations are determined manually. In order to annotate a question $q_i$, a human annotator must first manually search through the text corpus $\mathcal{D}$ in order to identify the document $d_i$ that answers the question. In a second step, a human annotator manually reads through the document $d_i$ and marks the answer span $s_i$. 

We assume separate costs, which are fixed over time, for every annotation-level. The price of annotating a document for a given question is defined as $\cdm$ and the price of annotating an answer span to a given question-document tuple as $\csm$. We explicitly distinguish these costs as the tasks can be of differing difficulty.

\textbf{Semi-supervised annotation (\smart) scheme:}
This scheme is supposed to reduce human effort by presenting candidates for annotation, so that only simple binary feedback is needed. In particular, human annotators no longer need to search through the entire document or corpus. Instead, we use the QA model $\Omega$ to generate a set of candidates (e.g., top-ranked documents or answer spans) and ask human annotators to give binary feedback in response (e.g., accept the candidate or reject it). This replaces the complex search task with a simpler form of interaction. As an example, to annotate the answer span for a question-document pair, the human annotator would not be required to read the entire document $d_i$, but only to determine which of the top-$n$ answers provided by $\Omega^{S}(\langle q_i,d_i\rangle)$ are correct. We assume \smart costs $\cds$ to annotate a document and $\css$ to annotate an answer span.

The \smart scheme should make annotations more straightforward, as providing binary feedback requires less time than reading through the texts. Hence, we assume that $\css<\csm$ and $\cds<\cdm$ hold. However, semi-supervised annotations might fail when none of the candidates is correct (i.e., the human annotators reject all candidates). In this case, our framework must revert to the \man procedure in order to obtain a valid annotation. As a consequence, the associated cost will increase to the accumulated cost for both the \smart and the \man schemes. 

Note that, no matter which scheme is chosen in practice, all annotations are confirmed by human annotators and our resulting dataset will be equal in quality to those resulting from traditional annotation.

\subsection{Annotation Costs}

Both annotation schemes, \man and \smart, incur different costs that further vary depending on whether annotation is provided at document level ($c^D$) or at answer span level ($c^S$). For annotating documents, the cost amounts to 
\begin{equation}
    c^{D}(a|q_i,d_i^{\ast}) = 
\begin{cases}
 c_a^D, \text{ if }a = 0 \text{ or }(a =  1\text{ and }\\
 \quad ~~~~~~~~~~~~~~~~~~~~ d_i^{\ast}\in\Omega^D(q_i)),\\
 \cdm + \cds, \text{ ~~~otherwise}\\
\end{cases}
\end{equation} 
where $a=\{0,1\}$ is the selected annotation scheme and $d_i^{\ast}$ is the ground-truth document annotation. Hence, $d_i^{\ast}\in\Omega^D(q_i)$ indicates the candidate set contains the ground-truth annotation and \smart is successful.

\noindent For annotating answer spans, the cost is given by
\begin{equation}
    c^{S}(a|\langle q_i,d_i\rangle,s_i^{\ast}) = 
\begin{cases}
 c_{a}^{S}, \text{ if }a = 1 \text{ or } (a = 0 \text{ and }\\
 \quad ~~~~~~~s_i^{\ast}\in\Omega^S(\langle q_i,d_i\rangle),\\
 \csm + \css, \text{ ~~~otherwise}.\\
\end{cases}
\end{equation} 
Alternatively, we can write the cost function as a matrix of annotation costs (Tbl.~\ref{tab:misclassification}). The diagonal entries reflect the costs paid for choosing the optimal scheme. The off-diagonals refer to the costs paid for a sub-optimal method (misclassification costs). 

\begin{table}[]
\footnotesize
\centering
\setlength{\tabcolsep}{4pt}
\begin{tabular}{@{}ccc@{}}\toprule
\multicolumn{3}{c}{\bfseries Annotation Costs for a Document $d$}                       \\ \midrule
\multicolumn{1}{l|}{}             & Cost-optimal:  \man & Cost-optimal: \smart  \\ 
\multicolumn{1}{l|}{Selected:  \man}  &     $\cdm$        &        $\cdm$      \\
\multicolumn{1}{l|}{Selected: \smart } &      $\cdm+\cds$       &     $\cds$         \\ \bottomrule
\toprule
\multicolumn{3}{c}{\bfseries Annotation Costs for an Answer Span $s$}                    \\ \midrule
\multicolumn{1}{l|}{}            & Cost-optimal: \man & Cost-optimal: \smart  \\ 
\multicolumn{1}{l|}{Selected: \man}  &     $\csm$           &          $\csm$         \\
\multicolumn{1}{l|}{Selected: \smart } &    $\csm+\css$          &      $\css$        \\ \bottomrule
\end{tabular}
\caption{Costs for annotating documents (top) and answer spans (bottom). The costs depend on the selected annotation scheme (rows) and the scheme that would have been cost-optimal (columns).}
\label{tab:misclassification}
\end{table}

\section{Learning a Cost-Effective Policy}
\label{sec:policy}

\subsection{Objective}

We aim to minimize overall annotation cost via 
\begin{equation}
    \sum_{i}  \mathbb{E}_{\pi^D,\pi^S}\left[c^{D}(a|q_i,d_i^{\ast}) + c^{S}(a| \langle q_i,d_i\rangle, s_i^{\ast})  \right].
\label{eqn:mini}
\end{equation}
It is important to see that the QA model and the policy model are intertwined, with both having an impact on Eq.~\ref{eqn:mini}. Updating the policy models learns the trade-off between \smart and \man annotations and, hence, directly minimizes the overall costs. Updating the QA model $\Omega$ increases the number of times suggested candidates are correct and, therefore, the fraction of successful \smart annotations. For instance, when adapting to a new domain, only a small fraction of suggested candidate annotations are correct, limiting the effectiveness of the \smart annotation. However, as we annotate more samples, we improve $\Omega$ and thus more suggested candidate annotations will be correct. For this, we later specify suitable updates for both the QA model and the policy model.

\subsection{Annotation Procedure and Learning}

Our framework proceeds according to these nine steps when annotating a question $q_i$ (see Alg.~\ref{alg:annotation_process}): First, we predict a number of top-$n$ documents that would be shown to annotators in the case of \smart annotation (line 2). Next, we decide upon the annotation scheme conditional on the prediction from the QA model (line 3) and, based on the selected scheme, request the ground-truth document annotation (line 4). After receiving the ground-truth document and observing the annotation costs, we update our policy network in line 5 (see Sec.~\ref{sec:policy-updates}). Next, we predict a number of top-$n$ answer span candidates for the question-document pair (line 6) and then decide upon the annotation scheme in line 7. After receiving the answer span annotation and observing a cost (line 8), we again update our policy model (line 9). Finally, we update the QA model with the newly annotated training sample in line 10 (see Sec.~\ref{sec:model-updates}). In practice, both policy updates and QA model updates (lines 5, 9, and 10) are invoked after a batch of questions is annotated. Furthermore, we initialize all models with an existing dataset (e.g., SQuAD).

\begin{algorithm}
\small
\SetKwInOut{Input}{Input}
\Input{list of questions $[q_1, \ldots, q_m]$ text corpus $\mathcal{D}$; QA model $\Omega$; policy models $\pi^D$ and $\pi^S$ }
\KwResult{annotated dataset $\{ \langle q_i,d_i,s_i \rangle\}_{0<i<m}$ }
 \While{$i\leq m$}{
  $[d^{(1)},\ldots,d^{(n)}] \leftarrow \Omega^D(q_i)$; predict top-$n$ documents\\
  $a \leftarrow \pi^D(q_i,[d^{(1)},\ldots,d^{(n)}])$; decide upon annotation scheme\\
  $d_i \leftarrow annotate(q_i|a)$; annotate document \\
  update $\pi^D$ w.r.t. the observed costs $c^{D}(a|q_i,d_i)$;\\
  $[s^{(1)},\ldots,s^{(n)}] \leftarrow \Omega^S(\langle q_i, d_i \rangle)$; predict top-$n$ answer candidates\\
  $a \leftarrow \pi^S(q_i,[s^{(1)},\ldots,s^{(n)}])$; decide upon annotation scheme\\
  $s_i \leftarrow annotate(\langle q_i,d_i \rangle|a)$; annotate answer span \\
  update $\pi^S$ w.r.t. the observed costs $c^{S}(a| \langle q_i,d_i \rangle , s_i)$;\\
  update QA model $\Omega$ with $\langle q_i,d_i,s_i \rangle$

 }
 \caption{High-Level Procedure of Annotation and Learning}
 \label{alg:annotation_process}
\end{algorithm}

\subsection{Policy Updates}
\label{sec:policy-updates}

Updating our policy model proceeds in three steps. (1)~We calculate whether the chosen action for past annotations was cost-optimal (i.e., whether the policy should have chosen the other scheme for annotation or not). (2)~We use this information to update the policy model with a supervised binary classification objective. This trains the policy to predict the probability of an annotation scheme given a new sample $p(a|x)$ without taking costs into account. (3)~We find a cost-sensitive decision threshold that chooses the optimal action with respect to the costs. All three steps are repeated after a full batch of samples has been annotated. 

Separating the policy update and the cost-sensitive decision threshold has several benefits. First, we know from cost-sensitive classification that we can calculate an optimal threshold point for ground-truth probabilities $p(a|x)$ \cite[c.f.][]{Kaufmann.2001,Ting.2000}. Therefore, we can focus our effort on determining probabilities as accurate as possible. Second, the decision threshold is calculated only from the costs $c_a^S$ and $c_a^D$ and, hence, if costs change, we do not need to re-estimate parameters but can directly adjust our policy.

\textbf{\underline{(1)~Finding the cost-optimal action:}}
In order to train the policy with a supervised update, we require labels for the cost-optimal annotation scheme for a given sample. If we choose the \smart annotation scheme, we immediately know whether the action was cost-optimal or not. This is due to the fact that, if the semi-supervised annotation fails, we have to switch to the \man scheme to receive an annotation and pay both costs. On the other hand, if we choose the \man scheme, we can observe the optimal action only after receiving the ground-truth annotation: We can then simply run the QA model and validate whether the annotation was contained within the top-$n$ candidates. If so, the \smart action would have been the better choice; otherwise, choosing \man would have been cost-optimal.

\textbf{\underline{(2)~Supervised model updates:}}
Our policy model is a neural network with parameters $\theta$ that predicts an annotation scheme for a given sample $x$, i.e., 
\begin{equation}
    p(a|x) = \text{NN}_\theta(x).
\end{equation}
Note that we dropped the $S$ and $D$ indices here as both policies differ only in the neural network architecture used. We can then simply train the policy with a supervised binary cross-entropy loss given the cost-optimal action that we calculated beforehand. Since the \smart scheme is often sub-optimal in the beginning, the training data is highly imbalanced. Therefore, we down-sample past annotations with a sampling ratio of $\alpha$ such that our training data is equally balanced. 

\textbf{\underline{(3)~Cost-sensitive decision threshold:}}
Choosing the annotation scheme with the highest probability does not take the actual costs into account. For instance, if \smart annotations are much cheaper than \man annotations, we want to choose the semi-supervised scheme even if its probability for success is low. More formally, we want to choose the annotation scheme $a$ that has the lowest expected cost $R(a|x)$, i.e., 
\begin{equation}
    R(a \,|\, x) = \sum_{a'} p(a' \,|\, x) \, c(a,a')
    \label{eqn:risk}
\end{equation}
where $c(a,a')$ is the annotation cost for choosing scheme $a$ when the optimal scheme was $a'$ (see Tab.~\ref{tab:misclassification}). Since we used down-sampling in our training, we have to calibrate the probabilities $p(a|x)$ with the sampling ratio $\alpha$ \cite{Pozzolo.2015}. Assuming the calibrated probabilities are accurate, there exists an optimal classification threshold $\beta$ \cite{Kaufmann.2001} that minimizes Eq.~\ref{eqn:risk} and Eq.~\ref{eqn:mini}. Therefore, we define our policy as follows: 
\begin{equation}
    \pi(a|x,\alpha,\beta) = 
\begin{cases}
&1 \text{, if }\frac{\alpha \, p(a=1|x)}{(\alpha-1) \, p(a=1|x) +1} \geq \beta \\
    &0 \text{, otherwise}.
\end{cases}
\label{eqn:policy}
\end{equation}
The optimal $\beta$ can be derived from the classification cost matrix \cite{Kaufmann.2001}  via
\begin{equation}
    \beta = \frac{c(1,0)-c(0,0)}{c(1,0)-c(0,0)+c(0,1)-c(1,1)},
\label{eqn:beta}
\end{equation}
again omitting superscripts $D$ and $S$ for brevity. Eq.~\ref{eqn:beta} is then simplified as the fraction of \smart annotation costs to \man annotation costs. Therefore, we can derive $\beta$ at document level
\begin{equation}
    \beta^D = \sfrac{\cds}{\cdm},
\end{equation}
and answer span level 
\begin{equation}
    \beta^S = \sfrac{\css}{\csm}.
\end{equation}

\subsection{Model Updates}
\label{sec:model-updates}

Periodically updating the QA model $\Omega$ allows the framework to adapt to the question style and domain at hand during the annotation process. Therefore, we improve the top-$n$ accuracy and the success rate of the \smart scheme over time. For practical reasons, we refrain from updating $\Omega$ after every annotation but periodically retrain the model after a batch of samples is annotated. In order to update the QA model, we can use the fully annotated QA samples in combination with a supervised objective. 

\section{Experimental Setup}

In this section, we introduce our experimental setup and implementation details.
\subsection{Datasets}
We base our experiments on the NaturalQuestion dataset \cite{Kwiatkowski.2019}. We choose this dataset as it is composed of about 300,000 real user questions posed to the Google search engine along with human-annotated documents and answer spans. Simulating the annotation of this dataset is similar to what would happen for domain customization of QA models in real practice (e.g., for search logs, FAQs, logs of past customer interactions). We focus on questions from the training-split that possess an answer span annotation and leave the handling of questions that do not have an answer for future work. The corpus for annotations is fixed to the English Wikipedia,\footnote{We extracted articles from the Wikipedia dump collected in October 2019, as this is close to the time period in which the NaturalQuestions dataset was constructed.} containing more than 5 million text documents.

\noindent \textit{Simulation of annotations:} Annotations in our experiments are simulated from the original dataset. If the framework chooses \man annotation, we simply use the original annotation from the dataset. If a \smart annotation is chosen, we simulate users that give positive feedback only to the ground-truth document and to the answer spans where the text matches\footnote{Here we only count exact matches.} the ground-truth annotation. We then construct the new annotation using the candidate with positive feedback. Since we simulate annotations, we conduct extensive experiments on how annotation costs influence the performance of our framework.

\subsection{Baselines}

To the best of our knowledge, there is no comparable prior work. Owing to this fact, we evaluate our framework against several customized baselines. First, we compare our approach against a manual annotation baseline in which we always invoke the full \man method to annotate samples. This represents the traditional method of annotating QA datasets and thus our prime baseline. Second, we draw upon a clairvoyant oracle policy that always knows the optimal annotation method. We use this baseline to report an upper bound of the savings that our framework could theoretically achieve. Third, we use our framework without updates on the QA model $\Omega$. This quantifies the cost-savings achieved by the interactive domain customization during annotation. Finally, we present a randomized baseline where the annotation scheme is decided by a randomized coin-toss.

\subsection{Implementation Details}

The QA model $\Omega$ is built as follows. We use a state-of-the-art BERT-based \cite{devlin2018bert} implementation of RankQA \cite{kratzwald2019rankqa}. This combines a simple tf-idf-based information retrieval module with BERT as a module for machine comprehension. Both policy models $\pi^D$ and $\pi^S$ are implemented as three-layer feed-forward networks with dropout, ReLu activation, and a single output unit with sigmoid activation in the last layer. For the policy $\pi^D$, we use the information retrieval scores as input. For $\pi^S$, we use the statistical features of answer-span candidates as calculated by RankQA \cite{kratzwald2019rankqa} as input. We also experimented with convolutional neural networks directly on top of the last layer of BERT, but without yielding improvements that justified the additional model complexity. We initialize all models with the SQuAD dataset (see our supplements).

\noindent \textit{Hyperparameter setting:} We set the number of candidates that are shown to annotators during a \smart annotation to $n=5$. The policy networks decide upon the annotation method based on features of the $2n$ highest-ranked candidates, i.e., the top-$10$. The batch size for updates in Alg.~\ref{alg:annotation_process} is set to 1,000 annotated questions. Details on hyperparameters of our QA and policy models are provided in the supplements.

\section{Experimental Results}

We group our experiments into three parts. First, we focus only on annotating the answer span for given question-document pairs, as this is the more challenging task.\footnote{Manually finding an answer span involves reading a document in depth. Manual document annotation is easier, as it can be supported with tools such as search engines. In such a case, our framework could still be used for answer span annotation, as is shown in Section \ref{ref:answerspanresult}.} Second, we carry out a sensitivity analysis in order to demonstrate how our framework adapts to different costs of \smart annotations and to show that we never exceed the cost of traditional annotation. Third, we evaluate our framework based on the annotation of a full dataset, including both answer span and document annotations, in order to quantify savings in practice by using our framework.

\subsection{Performance on Answer Span Annotations}
\label{ref:answerspanresult}
The annotation framework was used to annotate 45 batches of question-document pairs with the corresponding answer spans. The annotation costs are set to one price-unit for each \man annotation and one third of the unit for each \smart annotation. (In the next section, we carry out an extensive sensitivity analysis where the ratio for annotation costs between \man and \smart is varied.)

\begin{figure}[h]
    \centering
    \includegraphics[width=.98\linewidth]{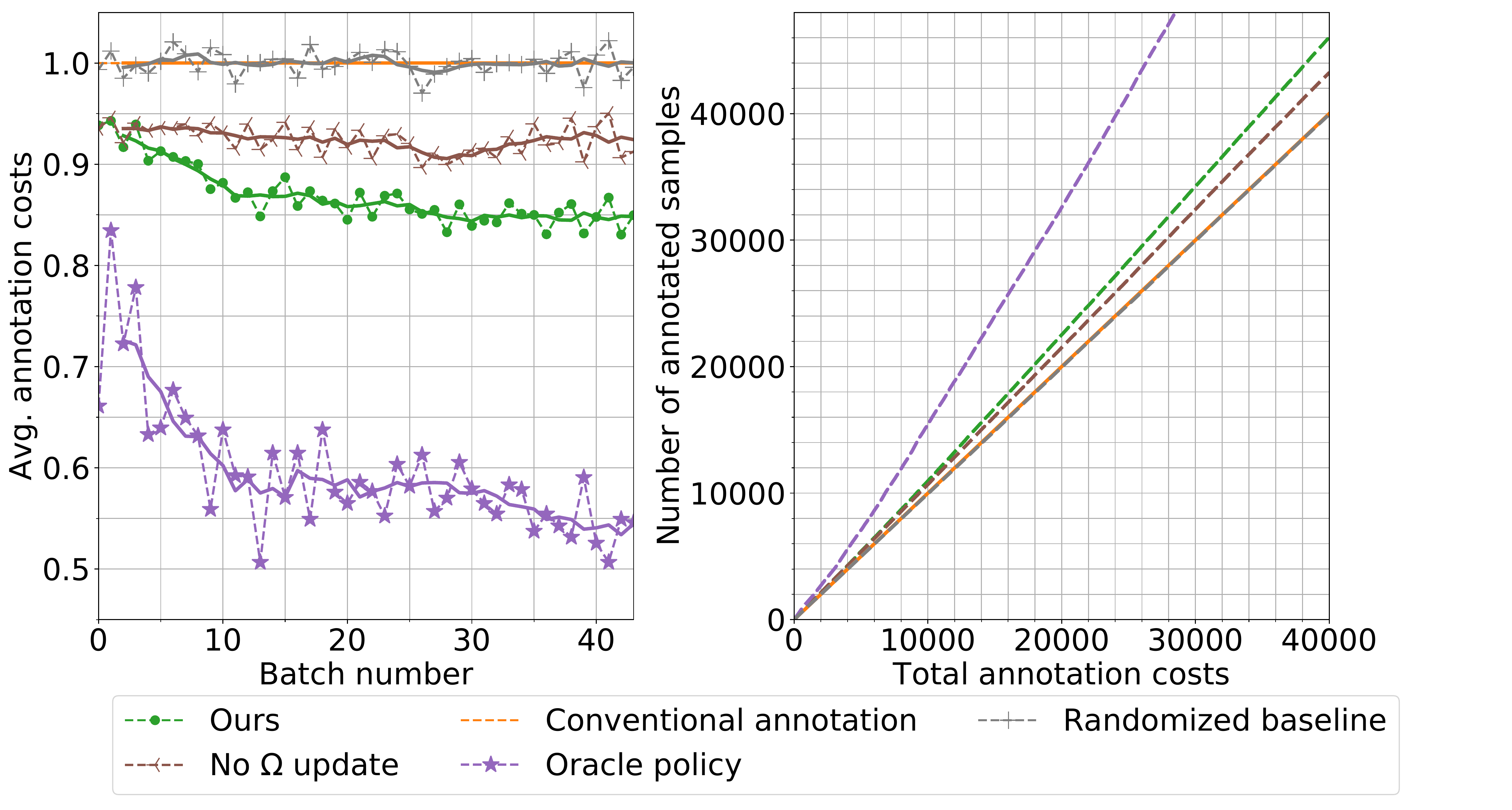}
    \caption{Left: average annotation costs in every batch as a dashed line, together with a running mean as a solid line. Right: how many samples we could annotate (y-axis) with a restricted available budget (x-axis).}
    \label{fig:main_plot}
\end{figure}

In Fig.~\ref{fig:main_plot} (left), we plot the average annotation costs in every batch with a dashed line, together with a running mean depicted as a solid line. Compared to conventional, manual annotation, our framework successfully reduces annotation cost by around 15\% after only 20 batches. We further compare it with an oracle policy that always picks the best annotation method. The latter provides a hypothetical upper bound according to which approximately $40$--$45\%$  of annotation cost could be saved. Finally, we show the performance of our framework without updates of the QA model $\Omega$. Here we can see that its improvement over time is lower, as the framework is not capable of adapting to the question style and domain used during annotation. In sum, our framework is highly effective in reducing annotation cost. 

Fig.~\ref{fig:main_plot} (right) shows how many samples we could annotate (y-axis) with a restricted budget (x-axis). For instance, assume we have a budget of 40k price units available for annotation. Conventional, manual annotation would result in exactly 40k annotated samples as we fixed the cost for each \man annotation to one unit. With the same budget, our annotation framework with semi-supervised annotations succeeds in annotating an additional $\sim$9,000 samples. 

\begin{table*}[h]
\centering
\footnotesize
\resizebox{\linewidth}{!}{
\begin{tabular}{@{}lrrrrrrrrr@{}}
\toprule
                       & \multicolumn{3}{c}{Document-level} & \multicolumn{3}{c}{Answer span-level}& \multicolumn{3}{c}{Overall} \\ 
                       \cmidrule(l){2-4}\cmidrule(l){5-7}\cmidrule(l){8-10}
 &$c_1=\sfrac{1}{4}$ & $c_1=\sfrac{1}{3}$ & $c_1=\sfrac{1}{2}$ &$c_1=\sfrac{1}{4}$ & $c_1=\sfrac{1}{3}$ & $c_1=\sfrac{1}{2}$ &$c_1=\sfrac{1}{4}$ & $c_1=\sfrac{1}{3}$ & $c_1=\sfrac{1}{2}$ \\ \midrule
Traditional Annotation & 102.4 & 102.4 & 102.4 &  102.4 & 102.4 & 102.4 & 204.8 & 204.8 & 204.8 \\
Ours & 79.8 & 85.6 & 98.0 & 81.8 & 87.0 & 98.3 & 161.6 & 172.7 & 196.3 \\
 & \scriptsize{(\textbf{22.1\%})} & \scriptsize{(\textbf{14.9\%})} &  \scriptsize{(\textbf{4.2\%})}&  \scriptsize{(\textbf{20.1\%})} &  \scriptsize{(\textbf{14.9\%})} &  \scriptsize{(\textbf{4.0\%})} &  \scriptsize{(\textbf{21.1\%})}  &   \scriptsize{(\textbf{15.7\%})}  &  \scriptsize{(\textbf{4.1\%})} \\
\bottomrule
\end{tabular}
}
\caption{Overall cost ($\times 10^3$ price unit) for annotating the NaturalQuestions dataset using our framework vs. conventional manual annotation for different \smart costs ($c_1$). Improvements are shown in parenthesis.}
\label{tab:overallannotation}
\end{table*}

\subsection{Cost-Performance Sensitivity Analysis}

\begin{figure}[h]
    \centering
    \includegraphics[width=.98\linewidth]{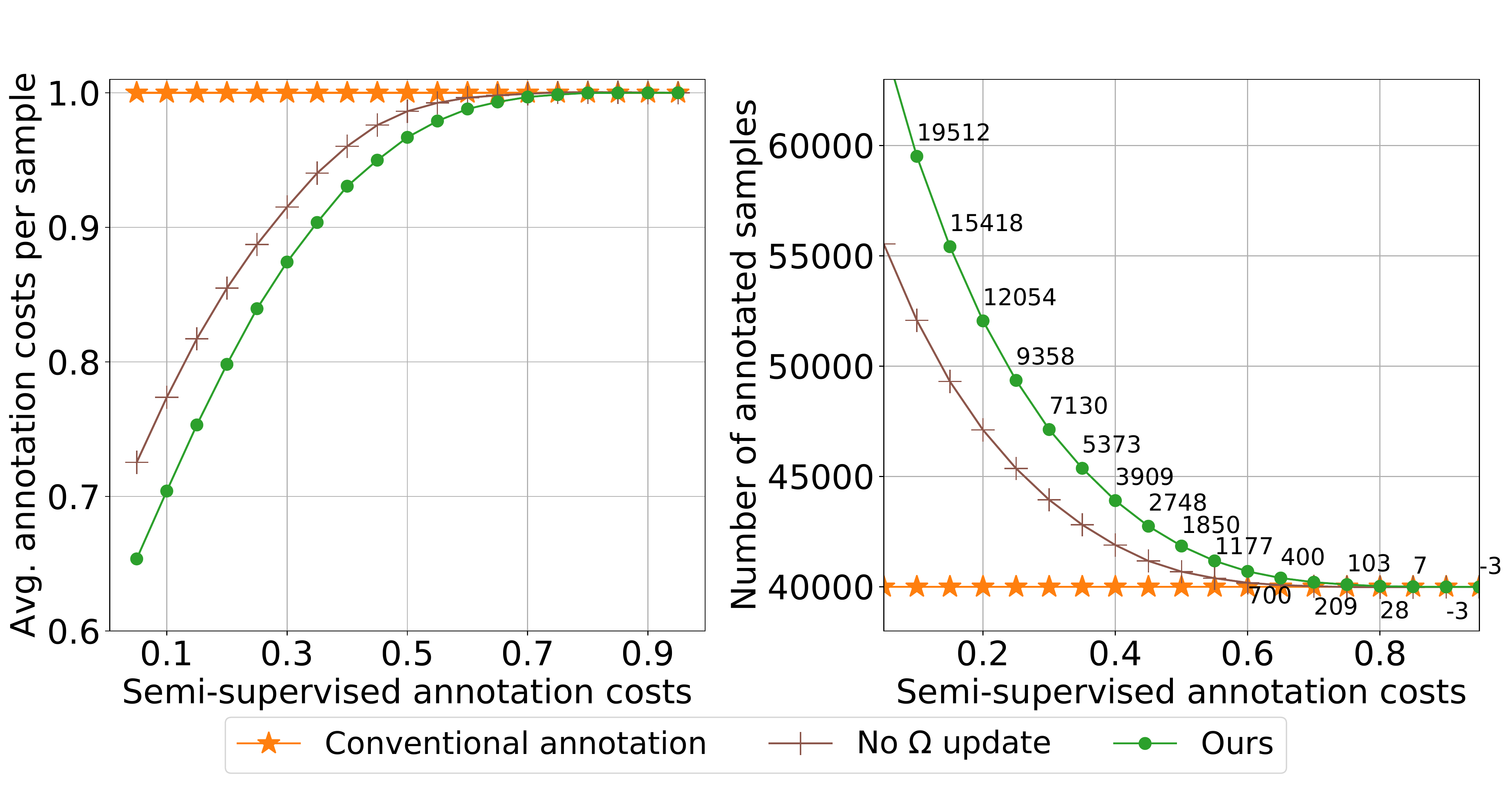}
    \caption{Left: average annotation costs when varying the ratio of \smart over \man annotation costs. Right: number of samples that were annotated with a restricted budget for a given \smart annotation cost.}
    \label{fig:robustness_plot}
\end{figure}

The advantage of our framework over manual annotations depends on the cost ratio between the \smart and \man schemes. In order to determine this, we identify the cost-range in which our framework is profitable as a function of \smart annotation costs. We study this via the following experiment: we repeatedly annotate 40k samples and keep the \man annotation costs fixed to one price unit, while we increase the costs of smart annotations from $0.05$ to $0.95$ in increments of $0.05$. Finally, we measure the average annotation costs for a single sample; see Fig.~\ref{fig:robustness_plot} (left). 

Fig.~\ref{fig:robustness_plot} (left) demonstrates that our framework effectively lowers annotation costs when the price for \smart annotations drops below $0.6$ as compared to manual annotations, which are fixed to one price-unit. Most notably, even when \smart annotations become expensive and almost equal the costs of \man annotations, the average annotation costs do not exceed those of strictly manual annotation. This can be attributed to our cost-sensitive decision threshold, which does not require exploration as in reinforcement learning, but directly sets the threshold in Eq.~\ref{eqn:policy} sufficiently high.

In Fig.~\ref{fig:robustness_plot} (right), we again show the number of samples that were annotated with a restricted budget of 40k price units. We marked the absolute gain in number of samples over traditional annotation in the plot. The benefit of our framework becomes evident once again when the ratio of \smart annotation costs to \man annotations costs falls below 0.6.  

To summarize, our framework is highly cost-effective: it reduces overall annotation costs or, alternatively, increases the number of annotated samples under a restricted budget if annotation costs of \smart are approximately half those of \man. If the costs are less than half those of \man annotation, the benefits are especially pronounced. Even if this assumption does not hold, our framework never exceeds the costs of manual annotation and never results in fewer annotated samples.

\subsection{Performance on Full Dataset Annotation}
In the last experiment, we simulate a complete annotation of the NaturalQuestions dataset, including annotations at both document level and answer span level. By annotating a complete dataset, we want to quantify the savings of our framework in practice. We again set the cost of each \man annotation to one price-unit and repeated the experiment three times by setting the \smart annotation cost ($c_1$) to one quarter, one third, and one half of the price unit. The results are shown in Tbl.~\ref{tab:overallannotation}. Depending on relative cost ratio $c_1$, we are able to save between $4.1\%$ and $21.1\%$ percent of the overall annotation cost. This amounts to a total of 40,000 to 8,000 price units.\footnote{In these experiments, we obtain the overall cost by directly adding the costs of the two levels. Note that the cost can be different for document and answer span annotations, and that in such cases, our framework can still save costs at each level as shown in the table, although we cannot directly add up the costs as an overall sum.}

\section{Discussion and {Future Work}}

We assume for the purposes of this study that questions have an answer span contained in a single document and leave an extension to multi-hop questions and unanswerable questions to future research. The robustness of our framework is demonstrated in an extensive set of simulations and experiments. We deliberately choose to leave experiments including real human annotators to future research for the following reason. Outcomes of such an experiment would be sensitive to the design of the user interface as well as the study design itself. In this paper, we want to put the emphasis on the methodological innovation of our framework and the novel annotation scheme itself.

On the other hand, experiments involving real users would provide valuable insights concerning the annotation costs and the quality of a dataset annotated with our method. Furthermore, it would be worth investigating how inter-annotator agreement or potential human biases manifest in traditional datasets as compared to those generated with our framework. 

\section{Conclusion}

We presented a novel annotation framework for question answering based on textual content, which learns a cost-effective policy to combine a manual annotation scheme with a semi-supervised annotation scheme. Our framework annotates {all} given questions {accurately} while limiting costs as much as possible. We show that our framework never incurs higher costs than traditional manual annotation. On the contrary, it achieves substantial savings. For example, it reduces the overall costs by about $4.1\%$  when \smart annotations cost about half of \man annotations. When that ratio is lowered to one fourth, our framework can reduce the total costs by up top $21.1\%$. We think that our framework could contribute to more accessible annotation of datasets in the future and possibly even be extended to other fields and applications in natural language processing.

\section*{Acknowledgments}
We would like to thank the anonymous reviewers for their helpful comments. We are also deeply grateful to Ziyu Yao for her valuable feedback and help in brainstorming and formalizing this idea. This research was partly supported by the Army Research Office under cooperative agreements W911NF-17-1-0412, and by NSF Grant IIS1815674 and NSF CAREER \#1942980. The views and conclusions contained herein are those of the authors and should not be interpreted as representing the official policies, either expressed or implied, of the Army Research Office or the U.S. Government. The U.S. Government is authorized to reproduce and distribute reprints for Government purposes notwithstanding any copyright notice herein.

\bibliographystyle{acl_natbib}

\section*{Appendix}

\appendix
\section{Source Code}

All source code is available from \href{https://github.com/bernhard2202/qa-annotation}{github.com/bernhard2202/qa-annotation}.

\section{Details on the QA Model}
We use the same hyperparemter configuration as reported in \citet{kratzwald2019rankqa} without further fine-tuning. The model was initialized by training on the training split of the SQuADv1.1. dataset \cite{Rajpurkar.2016}.

\section{Details on the Policy Model}
The policy models $\pi^D$ and $\pi^S$ are implemented as feed forward networks composed of a dense layer with $k$ output units and relu actvation, a dropout layer with dropout probability $z$, a second dense layer with $k/2$ outputs and relu activation, a dropout layer with dropout probability $z$, and a dense layer with a single output and sigmoid activation. 

\textbf{Initialization:} for the first batch of annotations we initialize the policy models on SQuAD. After the first batch is annotated we only use the new data for policy updates. 

\textbf{Hyperparameter search:} We tune hyperparamters on the SQuAD dataset using gridsearch with the values displayed in Tab.~\ref{tab:gridsearch} and Tab.~\ref{tab:gridsearch1}. Bold values mark final choices. We annotated the first 10 batches of SQuAD and choose the hyperparamters that had the lowest anntation cost. No hyperparemter tuning or architecture search was performed on the NaturalQuestions dataset which our experiments are based on.

\begin{table}[]
    \centering
    \begin{tabular}{c|c}
         Parameter &  Values\\ \hline
         Dropout z & 0.0, \textbf{0.3}, 0.5 \\
         Hidden units k & 32 64 \textbf{128} \\
         Learning rate & \textbf{0.0001}, 0.0005, 0.001 \\
         Epochs & 15, 20, \textbf{25}\\
    \end{tabular}
    \caption{Values used for gridsearch in hyperparameter tuning for the policy $\pi^S$ }
    \label{tab:gridsearch}
\end{table}

\begin{table}[]
    \centering
    \begin{tabular}{c|c}
         Parameter &  Values\\ \hline
         Dropout z & 0.0, \textbf{0.3}, 0.5 \\
         Hidden units k & 32 64 \textbf{128} \\
         Learning rate & \textbf{0.0001}, 0.0005, 0.001 \\
         Epochs & 15, 20, \textbf{25}\\
    \end{tabular}
    \caption{Values used for gridsearch in hyperparameter tuning for the policy $\pi^D$ }
    \label{tab:gridsearch1}
\end{table}

\section{Estimation of Real Annotation Costs}
In order to provide additional insights on the actual annotation costs involving real users we conducted a pre-test on Amazon MTURK. For this we showed a textual explanation of the \man and \smart annotation scheme to workers and provided them with mockups for both inputs (answer-span annotation). Next, we asked $40$ workers to report how much money they think would be a fair compensation for each of the tasks on a scale of one to ten. Workers reported on average a compensation of \$5.9 for \man annotations and \$3.2 for \smart annotations. This ratio falls into the range where we make profits using our framework. 

\section{System}
All experiments were conducted with a Nvidia Titan Xp GPU on a Server with 192GB DDR4 RAM and two 10 Core Intel Xeon Silver 4210 2.2GHz Processors.

\end{document}